\documentclass{article}

\usepackage[preprint]{neurips_2023}

\usepackage[utf8]{inputenc} % allow utf-8 input
\usepackage[T1]{fontenc}    % use 8-bit T1 fonts
\usepackage{hyperref}       % hyperlinks
\usepackage{url}            % simple URL typesetting
\usepackage{booktabs}       % professional-quality tables
\usepackage{amsfonts}       % blackboard math symbols
\usepackage{nicefrac}       % compact symbols for 1/2, etc.
\usepackage{microtype}      % microtypography
\usepackage{xcolor}         % colors
\usepackage{amsmath}
\usepackage{multirow}
\usepackage{graphicx}
\usepackage{gensymb}
\usepackage{courier}
\usepackage{natbib}
\usepackage{textcomp}
\usepackage{makecell}
\setcitestyle{numbers,square}

\title{Dive into the Power of Neuronal Heterogeneity}

\author{
	{Guobin Shen$^{1, 3}$, \ Dongcheng Zhao$^{1}$, \ Yiting Dong$^{1, 3}$, \ Yang Li$^{1, 4}$, \ Yi Zeng$^{1, 2, 3, 4}$}\thanks{Corresponding Author.}\\
	  $^1$ Brain-inspired Cognitive Intelligence Lab, Institute of Automation, Chinese Academy of Sciences\\ 
      $^2$ Center for Excellence in Brain Science and Intelligence Technology, CAS\\
      $^3$ School of Future Technology, University of Chinese Academy of Sciences \\
      $^4$ School of Artificial Intelligence, University of Chinese Academy of Sciences \\
	\texttt{\{shenguobin2021, zhaodongcheng2016, } \\ 
    \texttt{dongyiting2020, liyang2019, yi.zeng\}@ia.ac.cn}
}

\begin{document}

\maketitle

% TL;DR 在仅优化随机权重脉冲神经元的异质性参数时, 我们发现这样的异质性SNNs能够在强化学习, 工作记忆等任务上实现比同质性SNNs更好的性能, 并发现膜时间常数对于异质性SNNs至关重要, 并具有和生物神经元相似的分布. 

% Optimizing only heterogeneity parameters of random weighted SNN can improve performance in reinforcement learning and working memory. Membrane time constants are critical for heterogenous SNNs and exhibit a distribution similar to biological neurons.

% 生物神经网络是一个巨大的, 多样化的, 具有高度的神经多样性的结构. 通常的人工神经网络的只注重于通过训练修改连接的的权重, 神经元往往被建模成高度同质化的模型, 而缺乏对于神经异质性的探索. 仅有的一些关注神经元异质性的工作为了保证网络的性能, 会将神经元的属性和连接权重同时优化. 但这种策略也会对确定神经元异质性的具体贡献造成影响. 在本文中, 我们首先证明了基于反向传播的方法在对脉冲神经网络优化时面临的困境, 并使用了演化策略实现了对于随机网络中神经元参数的更稳健的优化. 在工作记忆, 机器人连续控制以及图像识别等任务上的实验表明, 神经元的异质性能够改善任务的表现, 特别是在长序列任务上. 此外, 我们发现膜时间常数在神经异质性中具有至关重要的作用, 并且其分布与生物实验中观察到的类似. 因此, 我们认为被忽视的神经异质性具有重要的作用, 这也为探索生物中神经元的异质性提供了新的手段, 同时也为设计更具生物合理性的神经网络提供了新的思路. 

\begin{abstract}
    The biological neural network is a vast and diverse structure with high neural heterogeneity. Conventional Artificial Neural Networks (ANNs) primarily focus on modifying the weights of connections through training while modeling neurons as highly homogenized entities and lacking exploration of neural heterogeneity. Only a few studies have addressed neural heterogeneity by optimizing neuronal properties and connection weights to ensure network performance. However, this strategy impact the specific contribution of neuronal heterogeneity. In this paper, we first demonstrate the challenges faced by backpropagation-based methods in optimizing Spiking Neural Networks (SNNs) and achieve more robust optimization of heterogeneous neurons in random networks using an Evolutionary Strategy (ES). Experiments on tasks such as working memory, continuous control, and image recognition show that neuronal heterogeneity can improve performance, particularly in long sequence tasks. Moreover, we find that membrane time constants play a crucial role in neural heterogeneity, and their distribution is similar to that observed in biological experiments. Therefore, we believe that the neglected neuronal heterogeneity plays an essential role, providing new approaches for exploring neural heterogeneity in biology and new ways for designing more biologically plausible neural networks.
\end{abstract}

\section{Introduction}

The biological neural network is an intricate system, and its high heterogeneity plays a vital role in learning and reasoning. In a biological neural network, the diversity of neurons endows the network with richer functionality and computational power. Different types of neurons can respond to different types of input features, thereby providing the network with better feature extraction capabilities~\cite{padmanabhan2010intrinsic, petitpre2018neuronal, lengler2013reliable}. The diversity of neurons enables the biological neural network to produce meaningful responses to external stimuli even without weight training. Furthermore, the heterogeneity of neurons is closely related to the plasticity of the brain and contributes to the robustness of neural networks~\cite{zeldenrust2021efficient, renart2003robust}.

However, conventional Artificial Neural Networks (ANNs) mainly rely on homogeneous activation functions to simulate how neurons receive information and adapt to different tasks through training the connection weights. In this process, neurons are often simplified and homogenized. In contrast, spiking neurons offer higher flexibility, and researchers can achieve different spiking patterns by adjusting various parameters~\cite{fangIncorporatingLearnableMembrane2021, shaban2021adaptive}. Nevertheless, these studies still cannot completely separate the influence of connection weights and independently optimize heterogeneous neurons. This could potentially obscure the specific contribution of neuronal heterogeneity to network performance. To address this challenge, we discuss the computational capabilities exhibited by neuronal heterogeneity when using a network without modifying the connections. Such research helps us better understand how neuronal heterogeneity independently influences and provides valuable insights for future network designs.

\begin{figure}
    \centering
    \includegraphics[width=1.\linewidth]{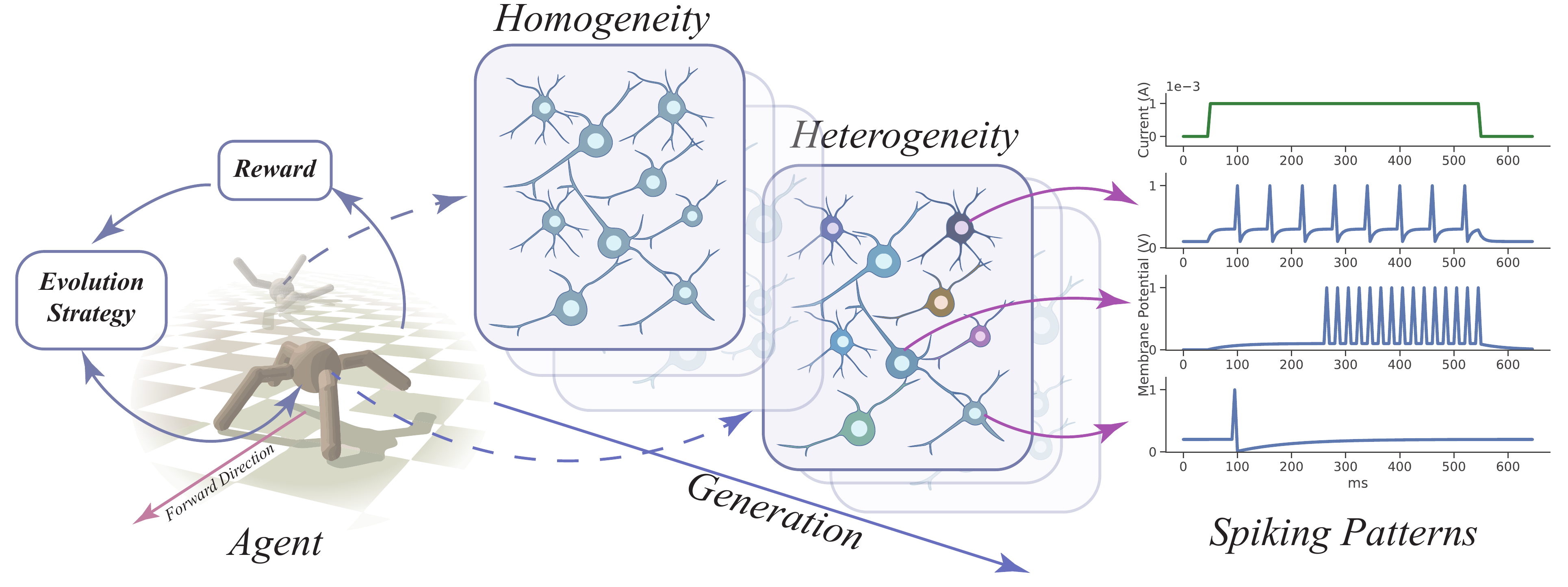}
    % 异质性神经网络的示意图. 左: 带有随机权重的异质性SNNs可以应用于机器人的连续控制任务. 右: 在输入电流恒定时, 异质性神经元能够展现不同的脉冲模式. 
    \caption{Diagram of Heterogeneous Neural Networks. Left: Heterogeneous SNNs with random weights can be applied to the continuous control tasks. Right: When the input current is constant, heterogeneous neurons can exhibit various spiking patterns.}
    \label{fig:intro}
\end{figure}

The success of deep learning in the field is largely attributed to the backpropagation (BP)~\cite{rumelhartLearningRepresentationsBackpropagating1986a}. However, there is still debate regarding whether the brain performs precise derivative computations~\cite{crick1989recent}. We first uncover the challenges faced by BP in optimizing Spiking Neural Networks (SNNs), particularly in terms of stability. Additionally, we discover that the Evolutionary Strategy (ES) without BP outperforms in optimizing the parameters of random networks, which aligns more closely with biological plausibility. Through experiments involving tasks like continuous control, working memory, and image recognition, we find that optimizing neuron properties not only allows neurons to exhibit diverse spiking patterns but also achieves performance that is comparable to, or even exceeds, that obtained by optimizing connection weights (as depicted in Fig.\ref{fig:intro}).

% 此外，我们的研究结果突出了膜时间常数在神经异质性中发挥的关键作用。有趣的是，我们观察到这些时间常数在我们模型中的分布与生物实验中的分布非常相似。这表明，被忽略神经元的异质性对于设计生物合理性的SNNs很重要，也为探索生物学中的神经异质性提供了新的机会。
Furthermore, our findings highlight the critical role of membrane time constants in neural heterogeneity. Interestingly, we observe that the distribution of these time constants in our model resembles that found in biological experiments. This suggests the significance of incorporating neglected neuronal heterogeneity in the design of biologically plausible SNNs and provides new opportunities for exploring neural heterogeneity in biology.

% 特别是，我们对探究神经元的异质性在SNNs中的作用有如下贡献.
In particular, our contributions to the exploration of the role of neuronal heterogeneity are as follows:

% 我们展示了基于反向传播的方法在优化脉冲神经网络时面临的困境, 尤其是在处理长序列任务时。我们发现使用无BP的演化算法对SNNs进行优化时, 能取得相比于BP更好的性能. 
% 我们利用一种进化算法来优化具有随机权重的网络中的神经元参数。实验结果表明，具有异质性神经元的随机权重网络能够达到与具有可优化权重的同质性网络相当或更高的性能。
% 通过对神经元不同特性的消融分析，我们强调了膜时间常数在神经异质性中的关键作用，以及它们与生物实验的分布相似性。
% \begin{itemize}
%     \item We demonstrated the challenges faced by the backpropagation-based approach in optimizing SNNs, particularly in handling long sequence tasks. We found that using BP-free evolutionary algorithms for optimizing SNNs can achieve better performance.
%     \item We utilize an evolutionary algorithm to optimize neuronal properties in networks with randomized weights. The experimental results show that optimizing neuron properties can achieve comparable or superior performance to optimizing connection weights, as shown in Fig.~\ref{fig:intro}.
%     \item Through the ablation analysis of the different properties of neurons, we highlights the crucial role of membrane time constants in neural heterogeneity, and their distribution similarities with biological experiments.
% \end{itemize}

\begin{itemize}
    \item We highlight the challenges of BP in optimizing SNNs, particularly in long sequence tasks, and show that BP-free evolutionary algorithms can yield better results.
    \item We employ ES to optimize the properties of neurons in a network with random weights. Experimental results demonstrate that the random weight networks with heterogeneous neurons can achieve comparable or even higher performance than the network with homogeneous neurons and trainable weights.
    \item Our ablation analysis underscores the importance of membrane time constants in neural heterogeneity and their distribution similarities with biological experiments.
\end{itemize}

% ---------------------- Removed for shorten ----------------------
% 总之，我们的工作有助于加深对生物和人工神经网络中神经异质性重要性的理解。通过引入无BP的方法和确定膜时间常数的关键作用，我们希望能激发新的研究途径和开发更多的生物学上合理的神经网络。通过对神经异质性的探索和整合，研究人员可能会释放出新的潜力来提高ANN的性能，并获得对生物神经网络的复杂世界的宝贵见解。
% In conclusion, our work contributes to a deeper understanding of the importance of neural heterogeneity in both biological and SNNs. By introducing a BP-free approach and identifying the key role of membrane time constants, we hope to inspire new avenues of research and the development of more biologically plausible neural networks. Through the exploration and incorporation of neural heterogeneity, researchers may unlock new potential for improved performance in SNNs and gain valuable insights into the complex world of biological neural networks.

\section{Related Works}

\subsection{Neural heterogeneity}

Recent studies highlight the critical role of heterogeneous neurons in enhancing the robustness and performance of SNNs~\cite{sheHeterogeneousSpikingNeural2021, poolSpikeTimingDependentPlasticityReliability2011}. This heterogeneity extends to neuron connectivity, synaptic properties, learning rules, and characteristics. For instance, the use of Hebbian learning rules for dynamic weight optimization allows networks to better adapt to unforeseen damages, enhancing their resilience and performance~\cite{najarroMetaLearningHebbianPlasticity2020}. Similarly, dynamic neuron heterogeneity can distinguish temporal patterns~\cite{sheSequenceApproximationUsing2022}, while the Heterogeneous Recurrent SNN (HRSNN) outperforms homogeneous SNNs in video activity recognition and classification~\cite{chakrabortyHeterogeneousNeuronalSynaptic}.

Neuronal properties, such as membrane time constants, greatly influence brain dynamics and functions~\cite{decoBrainSongsFramework2019a, mattiaPopulationDynamicsInteracting2002a, hasselmoMechanismsUnderlyingWorking2006}. Despite this, few studies have modeled neuron heterogeneity in SNNs. Such modeling can enhance SNNs' performance in real-time tasks~\cite{perez-nievesNeuralHeterogeneityPromotes2021} and accelerate learning~\cite{fangIncorporatingLearnableMembrane2021}. However, optimization of both synaptic weights and neuronal properties is challenging and makes it hard to determine the specific impact of neuronal heterogeneity. Previous research has concentrated on analyzing membrane time constants, often overlooking other spiking neurons properties like threshold voltage and resting potential. In response to these gaps in the existing literature, our work focuses on investigating the importance of heterogeneous neurons in a network with random weights, and the distinct contributions of different neuronal properties.

\subsection{Randomized Networks}

Weight Agnostic Neural Networks (WANNs), first introduced by Gaier and Ha~\cite{gaierWeightAgnosticNeural2019}, deviate from conventional deep learning by prioritizing architecture over weight fine-tuning. This approach allows WANNs to perform tasks effectively with random or shared weights, resulting in efficient and lightweight models. Network training is achieved by adjusting the architecture and neuron activation functions, employing a minimalistic algorithm, NeuroEvolution of Augmenting Topologies (NEAT)~\cite{stanleyEvolvingNeuralNetworks2002}, for evolving network architectures.

Despite using suboptimal weights, WANNs perform competitively on tasks like CartPole, bipedal locomotion, and MNIST~\cite{lecun1998mnist}. However, their use of continuous non-linear activation functions and limited weight evaluation per structure can increase training costs, posing a challenge for efficient learning.

Following the pioneering work of Gaier and Ha~\cite{gaierWeightAgnosticNeural2019}, we examine the role of neuron heterogeneity in fixed, randomly initialized SNNs without the influence of other optimizable parameters.

% TOOD: Continue Check

\section{Method}

% 异质性的神经元, 公式以及可训练参数;
% BP遇到的问题, 公式 
% 我们用 ES + Random Network

\subsection{Heterogeneous Spiking Neuron}

% LIF（Leaky Integrate-and-Fire）神经元是一种简化的神经元模型，用于研究神经网络的动力学行为。它是从生物神经元的行为中抽象出来的，通过一个相对简单的数学模型来描述神经元的充电和放电过程。LIF神经元模型在计算神经科学中被广泛应用，因为它在捕捉神经元基本动力学特性和计算复杂性之间取得了一个很好的平衡。LIF神经元模型的基本公式可以表示为：

The Leaky Integrate-and-Fire (LIF) neuron is a simplified neuron model used to study the dynamic behavior of neural networks. It is an abstraction from the behavior of biological neurons, describing neurons' charging and discharging process through a relatively simple mathematical model. The LIF neuron model is widely used in computational neuroscience because it strikes a good balance between capturing the essential dynamic characteristics of neurons and computational complexity. The basic formula of the LIF neuron model can be expressed as follows:
\begin{equation}
    \tau_m \frac{\partial v}{\partial t} = -(v - v_{rest}) + RI(t)
    \label{lif_continuous}
\end{equation}

% 在式1中, %\tau_m% 是膜时间常数, 决定神经元充电和放电过程的速度. $I$是输入电流, $v_{rest}$ 是静息电位, 决定了神经元在没有输入电流时候的膜电位, $R$ 是膜电阻, $v_{th}$ 是阈值电位. 当膜电位达到阈值电位时，神经元会发放一个动作电位, 膜电位会恢复为静息电位. 为了进行数值模拟，通常需要对LIF神经元模型进行离散化。采用欧拉方法对膜电位方程进行离散化：

In Eq.~\ref{lif_continuous}, $\tau_m$ is the membrane time constant, determining the speed of the neuron's charging and discharging processes. $I(t)$ represents the input current, and $v_{rest}$ is the resting potential, which determines the neuron's membrane potential when there is no input current. $R$ is the membrane resistance, and $v_{th}$ is the threshold potential. When the membrane potential reaches $v_{th}$, the neuron fires a spike, and the membrane potential returns to the resting potential. It is often necessary to discretize the LIF neuron model to perform numerical simulations. The neuron dynamics equation can be discretized using the Euler method:
\begin{equation}
    \begin{aligned}
        u(t) & = v(t - \Delta t) + \frac{\Delta t}{\tau_m}(R\sum_i w_i s_i(t) - v(t - \Delta t) + v_{rest}) \\
        s(t) & = g(u(t) - v_{th})                                                                           \\
        v(t) & =  u(t) (1 - s(t)) + v_{reset} s(t)                                                          \\
        \label{lif}
    \end{aligned}
\end{equation}

In Eq.~\ref{lif}, $\Delta t$ is the time step, $v_{reset}$ is the reset potential, $u(t)$ and $v(t)$ represent pre- and post-spike membrane potentials, and $g(\cdot)$ is the Heaviside function modeling spiking behavior. Synaptic weights, represented as $w_i$, are often prioritized in conventional neural networks. Despite this, our study focuses on the overlooked aspect of neuronal heterogeneity in randomly weighted SNNs, aiming to optimize neuron parameters like membrane time constant, membrane resistance, resting potential, and threshold voltage. In the experimental section, we make $v_{reset} = v_{rest}$ for simplicity.

\subsection{Spiking Neurons: Iterative Dynamical System}
\label{sec:dynamic_system}

% LIF神经元可以被认为是一个迭代的动态系统. 不失一般性的, 一个具有有限步数$N$的脉冲神经元的状态转移方程可以被描述为作用于脉冲后膜电位, 输入电流和神经元属性的非线性函数$f$:
% A LIF neuron can be considered as an iterative dynamic system. Without loss of generality, the state transition equation for a spiking neuron with a finite number of steps $N$ can be described as a nonlinear function $f$ acting on the post-spike membrane potential $v$, input current $I$, and neuron properties $\theta = \{\tau_m, v_{th}, v_{rest}, R\}$:
% \begin{equation}
%     u(t + \Delta t) = f(u(t), I(t + \Delta t); \theta) 
%     \label{generality}
% \end{equation}

% 在控制理论和强化学习中，我们通常希望优化控制变量在一个轨迹上的某个目标函数。例如，我们可能希望找到一种策略，使得在有限时间内，期望总成本最小或期望总奖励最大。为了评估策略的质量，通常定义一个损失函数，该函数衡量策略在有限步数N内的表现。例如，再以最后一层神经元的膜电位作为网络的输出时, 考虑以下损失函数，它对计算到有限步数N的损失（lt）进行求和：

In control theory and reinforcement learning, the primary objective is to optimize a control variable towards a specific objective function along a trajectory. For instance, we aim to find a policy that minimizes the expected total cost or maximizes the expected total reward within a finite time horizon. To evaluate the quality of a policy, it is common to define a loss function that measures the performance of the policy over a finite number of steps $N$. For example, when considering the output of a neural network as the membrane potential of its last layer, we can define the following loss function, which sums up the losses $L_n$ computed up to a finite step $N$.

\begin{equation}
    L(\theta) = \frac{1}{N}\sum_{n=0}^{N-1} l_n(v(n \Delta t); \theta)
    \label{loss_fn}
\end{equation}

% 在式~\ref{loss_fn}中, $\theta$ 表示可训练的参数, 在本文中, 可训练参数是神经元的参数, 也即$\theta = \{\tau_m, v_{th}, v_{rest}, R\}$. 对于基于梯度的优化方法而言, 损失函数的导数常被关注, $t$ 时刻损失函数的导数可以被表示为: 
In Eq.~\ref{loss_fn}, $\theta$ denotes the set of trainable parameters. In this context, the trainable parameters are the neuron parameters, i.e., $\theta = \{\tau_m, v_{th}, v_{rest}, R\}$. For gradient-based optimization methods, the derivative of the loss function is of particular interest. The derivative of the loss function at time $t$ can be expressed as:
\begin{equation}
    \begin{aligned}
        \frac{\partial L_n}{\partial \theta} & = \frac{\partial l_t}{\partial v(t)} \frac{\partial v(t)}{\partial \theta} + \frac{\partial l_t}{\partial v(t)} \frac{\partial v(t)}{\partial v(t - \Delta t)} \frac{\partial v(t - \Delta t)}{\partial \theta} + \dots + \frac{\partial l_t}{\partial v(t)}  \dots \frac{\partial v(\Delta t)}{\partial v(0)} \frac{\partial v(0)}{\partial \theta} \\
                                             & = \sum_{k=0}^{n} \frac{\partial l_n}{\partial v(t)} (\prod_{i=k+1}^{n} \frac{\partial v(i\Delta t)}{\partial v((i-1)\Delta t)}) \frac{\partial v(k\Delta t)}{\partial \theta}
    \end{aligned}
    \label{loss_step}
\end{equation}

% 根据Eq.3, 损失对于x的总的梯度可以被表示为

According to Eq.~\ref{loss_fn}, the total gradient of $L(\theta)$ with respect to $\theta$ is:
\begin{equation}
    \nabla_{\theta} L(\theta) = \frac{1}{N} \sum_{n=0}^{N}[\sum_{k=0}^{n} \frac{\partial l_n}{\partial v(t)} (\prod_{i=k+1}^{n} \frac{\partial v(i\Delta t)}{\partial v((i-1)\Delta t)}) \frac{\partial v(k\Delta t)}{\partial \theta}]
    \label{loss_total}
\end{equation}

% 具体的对于不同神经元的参数的损失可以在补充材料xxx中找到. 在式5中, 出现了乘, 其中的元素正是LIF神经元的膜电位转移方程的Jacobian矩阵. 根据公式2, 可以给出其递推形式.
% Specific loss functions for different neuron parameters can be found in the Supplementary Material (TODO). 
Eq.~\ref{loss_total} involves a cumulative product of the Jacobian matrix, which corresponds to the membrane potential transfer function of the LIF neuron. According to Eq.~\ref{lif}, its recursive form can be derived.
\begin{equation}
    \frac{\partial v(t)}{\partial v(t - \Delta t)} = [1 - s(t) + (v_{rest} - u(t)) g^\prime (u(t) - v_{th})](1 - \frac{\Delta t
    }{\tau_m})
    \label{recurisve}
\end{equation}

% \begin{equation}
%     \begin{aligned}
%          \frac{\partial v(t)}{\partial v(t - \Delta t)} &=  \frac{\partial v(t)}{\partial u(t)} \frac{\partial u(t)}{\partial v(t - \Delta t)} + \frac{\partial v(t)}{\partial s(t)}\frac{\partial s(t)}{\partial u(t)} \frac{\partial u(t)}{\partial v(t - \Delta t)} \\
%         &= [1 - s(t)](1 - \frac{\Delta t}{\tau_m}) + [-u(t)+v_{reset}]g^\prime (u(t) - v_{th})(1 - \frac{\Delta t}{\tau_m}) \\
%         &= [1-s(t)+(-u(t)+v_{rest})g^\prime (u(t) - v_{th})]((1 - \frac{\Delta t}{\tau_m})
%     \end{aligned}
%     \label{loss_step}
% \end{equation}

% 在式xx中, $g^\prime$ 表示heaviside函数的导数, 但是由于其在不可微分, 因此经常使用其近似的形式, 这被称为代理函数. 代理梯度在赋予了SNNs通过反向传播优化的便利的同时, 也使得梯度的估计变得不准确. 当式5中xxx的一些或者所有的特征值的大小大于$1$, 那么系统就会发散. 而$u(t)$ 则是脉冲前的膜电位, 一般被认为是无界的, 并在实践中发现其绝对值可以变得很大. 这就难以保证在使用反向传播对SNNs进行优化时, 特别是对于长序列任务时候的动力学系统的稳定性. 为了缓解这一问题, 一些研究在实际应用时, 会断开$v(t)$ 和 $s(t)$ 的梯度链路, 以保证式子6的大小小于$1$, 进而保证系统的稳定. 但是这样做却也让时间维度的梯度呈现指数衰减, 限制了SNNs的长时记忆能力. 更详细的关于SNNs稳定性的讨论请参考附录A.1.
In Eq.~\ref{recurisve}, $g^\prime$ represents the derivative of the Heaviside function. Due to its non-differentiability, an approximate form is often used, known as a surrogate function~\cite{bohte2011error}. The surrogate gradient provides the convenience of optimizing SNNs through backpropagation but also introduces inaccuracies in gradient estimation. When some or all of the eigenvalues in $\prod_{i=k+1}^{n} (\frac{\partial v(i\Delta t)}{\partial v((i-1)\Delta t)})$ of Eq.~\ref{loss_total} are larger than $1$, the system may diverge~\cite{bollt2000controlling}. Additionally, $u(t)$ represents the membrane potential before the spike, which is generally considered unbounded and can have large absolute values in practice~\cite{guo2022recdis}. This makes it challenging to ensure the stability of the dynamical system, especially for long sequence tasks, when using backpropagation to optimize SNNs. To mitigate this issue, some studies disconnect the gradient flow for $v(t)$ and $s(t)$ in applications~\cite{wu2018spatio} to ensure that the magnitude of Eq.~\ref{recurisve} is smaller than $1$, thereby ensuring system stability. However, this approach also results in exponentially decaying gradients along the time dimension, limiting the long-term memory capacity of SNNs.

% 在公式\ref{recurisve}中，$g^\prime$是Heaviside函数的导数。它的非微分性往往需要一个近似的代用函数~\cite{bohte2011error}，通过BP帮助SNN优化。然而，潜在的系统发散和大的尖峰前膜电位绝对值，$u(t)$~cite{guo2022recdis}，使得稳定性受到挑战，特别是对于长序列任务。一些研究将$v(t)$和$s(t)$的梯度流断开以确保稳定性~cite{wu2018spatio}，但这导致梯度的指数衰减，限制了SNNs的长期记忆能力。
% In Eq.\ref{recurisve}, $g^\prime$ is the derivative of the Heaviside function. Its non-differentiability often necessitates an approximate surrogate function~\cite{bohte2011error}, aiding SNN optimization through BP. However, potential system divergence and large absolute values of pre-spike membrane potential, $u(t)$~\cite{guo2022recdis}, make stability challenging, especially for long sequence tasks. Some studies disconnect the gradient flow for $v(t)$ and $s(t)$ to ensure stability~\cite{wu2018spatio}, but this results in exponentially decaying gradients, limiting SNNs' long-term memory capacity.

\subsection{Evolutionary Strategy}

% 一个可行的方法是丢掉所有的梯度, 采用一些黑箱的方法来估计优化的方向. 这也在一定程度上缓解了SNNs在使用BPTT时候, 需要存储所有的轨迹, 造成的存储开销随序列长度线性增长的问题~\cite{yu2018sliced}. 因此我们尝试使用BP-free的演化方法对于神经元的异质性进行研究, 并与基于反向传播的方法进行对比. 
% One feasible approach is to discard all gradients and adopt some black-box methods to estimate the optimization direction. This also alleviates the storage cost issue caused by storing all trajectories when using BPTT in SNNs, which increases linearly with sequence length ~\cite{yu2018sliced}. Therefore, we attempt to study the heterogeneity of neurons using BP-free evolutionary methods and compare it with the gradient-based methods.

A viable approach discards gradients and uses black-box methods to estimate the optimization direction, alleviating the storage cost issue of storing all trajectories in BP Through Time~(BPTT)~\cite{yu2018sliced}. We explore neuronal heterogeneity using BP-free evolutionary methods, contrasting with gradient-based methods.

% 具体来说, 我们通过添加扰动的方式, 对梯度进行更鲁棒的估计. 假设扰动的标准差是$\sigma$, 初始的神经元参数是$\theta_0$, 并在标准正态分布中采样$M$个扰动$\epsilon_i \sim \mathcal{N}(0, I)$. 那么根据扰动对于损失的梯度的估计是:

Specifically, we estimate the gradient more robustly by adding perturbations. Assuming the standard deviation of the perturbation is $\sigma$, the initial neuron parameters are $\theta_0$, and $M$ perturbations $\epsilon_i \sim \mathcal{N}(0, I)$ are sampled from a standard normal distribution. Then, the estimate of the gradient of the loss with respect to the perturbation is:
\begin{equation}
    \nabla_{\theta} L_{ES}(\theta) = \mathbb{E}_{\epsilon \sim \mathcal{N}(0, I)}\{\frac{\epsilon}{\sigma}L(\theta + \sigma \epsilon)\} = \mathbb{E}_{\epsilon \sim \mathcal{N}(0, I)}\{\frac{\epsilon}{\sigma}(L(\theta + \sigma \epsilon) - L(\theta)) \}
    \label{perturbation}
\end{equation}

% 那么根据Eq.7, 在学习率为$\alpha$时, 对于神经元参数的更新可以被表示为:
According to Eq.~\ref{perturbation}, the update of the neuron parameters can be expressed as follows, given a learning rate of $\alpha$:
\begin{equation}
    \theta_{t+1} \gets \theta_{t} + \frac{\alpha}{\sigma M}\sum_{j=0}^{M} L(\theta_t + \sigma \epsilon_j)
    \label{update_es}
\end{equation}

% 演化算法采用随机搜索策略对神经网络的参数进行优化, 这种方法与生物系统中自然选择的概念更为接近, 而且BPTT算法需要在时间上反向传播误差, 这在生物神经系统中是不现实的. 在实践中, 梯度的方差可能以指数形式增长, 在连续控制的背景下, 通过Eq.~\ref{perturbation}这种方式估计梯度, 能产生更低的估计误差~\cite{parmas2018pipps}. 
% Evolutionary algorithms use a random search strategy to optimize the parameters of neural networks, which is closer to the concept of natural selection in biological systems. In contrast, BPTT algorithms require the backward propagation of errors in time, which is not realistic in the biological neural system. In practice, the variance of gradients may increase exponentially. Estimating gradients by Eq.~\ref{perturbation} can produce lower estimation errors, particularly in the context of continuous control\cite{parmas2018pipps}.

Like natural selection in biological systems, ES uses random search strategies to optimize neural network parameters. These strategies can reduce estimation errors, especially in continuous control contexts~\cite{parmas2018pipps}, compared to BPTT.

\section{Result}

% 在这小节, 我们首先我们在工作记忆和机器人连续控制任务上, 对比了优化同质化神经网络中的连接权重和异质性神经网络中的神经元属性数对于性能的影响. 之后我们对于神经元的不同属性进行消融分析, 并将生物神经元的膜时间常数和SNNs中的时间常数的分布进行了对比. 随后, 我们在图像分类任务上对于随机权重的异质性SNNs进行了测试, 以说明异质性神经元的通用性.  最后我们在gymnax中一些经典的控制任务中对比了基于BP和ES算法在对于连接权重和神经元异质性进行优化时的性能. 实验中使用的网络结构, 神经元的和其他超参数的设置, 和训练方法可以在附录A.1中找到. 
In this section, we first compare the impact of optimizing connection weights in homogeneous neural networks and optimizing neuron properties in heterogeneous neural networks on working memory and continuous control tasks while keeping the same number of trainable parameters. We then conduct ablation analysis on different properties of neurons and compare the distribution of membrane time constants in biological neurons and SNNs. Afterward, we test heterogeneous SNNs with random weights on image classification tasks to demonstrate the generality of heterogeneous neurons. Lastly, we compare the performance of BP and ES in optimizing connection weights and neuron heterogeneity in some classic control tasks of the Gymnax environment~\cite{gymnax2022github}. Details regarding the network architecture, neuron parameters, hyperparameter settings, and training methods can be found in Tab.~\ref{params}.

\subsection{Neuronal Heterogeneity on Memory Tasks}

% 工作记忆是我们的认知系统的一个关键部分，它允许我们保持和处理短期的、正在进行的任务相关信息。对于一个合格学习系统而言, 工作记忆是不可或缺的. 而神经元的异质性可能是记忆的重要的神经元基础. 因此, 我们在bsuite中记忆长度任务中, 对于这一想法进行了验证. 该环境的底层是一个风格化的T形迷宫. 每个回合共有N个步骤, 每个时刻的状态可以被表示为 $s_t=(c_t, \frac{t}{N}$, 对于$\forall t = 0, \dots, N-1$. 动作空间为$A = \{-1, +1\}$. 其中, $c_0 \sim Unif(A)$, $c_t = 0$ 对于所有的 $t \geq 1$. 记忆任务的奖励是 $r_{N-1} = Sign(a_{N-1} = c_0)$.  该任务要求智能体记忆初始的刺激, 并在最后一步复现这一刺激. 对不同的实验设置, 我们使用不同的随机种子重复了50次, 并在不同长度的记忆任务上随机权重的异质性SNNs和权重可训练的SNNs进行了实验. 

Working memory constitutes a pivotal component of a cognitive system, allowing us to hold and manipulate task-relevant information in the short term. Neuronal heterogeneity potentially serves as a critical neural foundation for memory. Therefore, we validate this hypothesis in the \texttt{memory length} task within bsuite~\cite{osband2019behaviour}. This environment is a stylized T-maze~\cite{o1971hippocampus}, consisting of $N$ steps per episode. Each state can be represented as $s_t=(c_t, \frac{t}{N})$, for all $ t = 0, \dots, N-1$. The action space is $\mathcal{A} = \{-1, +1\}$, where $c_0 \sim Unif(A)$ and $c_t = 0$ for all $t \geq 1$. The reward for the memory task is $r_{N-1} = Sign(a_{N-1} = c_0)$. The task requires the agent to remember and reproduce the initial stimulus in the final step. We conduct experiments on randomly-weighted SNNs with heterogeneous neurons and weight-trainable SNNs for diverse experimental settings, using different random seeds and repeating the experiments $50$ times of varying lengths.

\begin{figure}
    \centering
    \includegraphics[width=1.\linewidth]{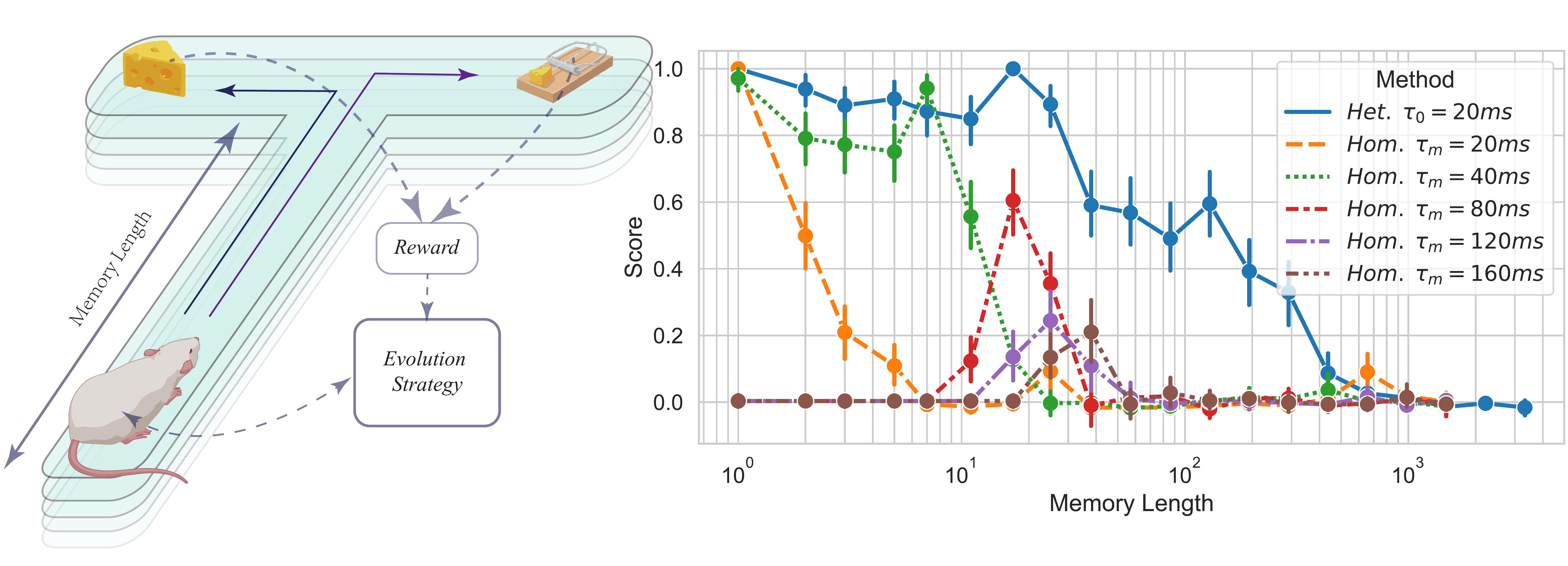}
    % 左: \texttt{memory length}环境的示意图. 右: 在不同的记忆步数时, 只优化神经元属性和只优化连接权重时, SNNs的性能. 
    \caption{Left: Illustration of the \texttt{memory length} environment. Right: Performance of SNNs when optimizing only neuron properties and only connection weights at different memory steps.}
    \label{fig:work_mem}
\end{figure}

% 如图xxx所示，具有异质性神经元的SNNs, 即使在突触权重不可优化的情况下，仍然实现了远超于传统SNN的记忆能力。随着任务难度的增大，异质性的SNNs在延迟为25步时仍能保持约0.9的奖励。相比之下，同质性的SNNs虽然可以通过调整连接权重适应任务，但是随着记忆长度的增加，其性能急剧下降。同时为了确保结论的鲁棒性, 我们对于同质化SNNs的膜时间常数进行了超参数搜索, 如Fig~\ref{fig:work_mem}中的右图所示, 对于同质化 SNNs，不同的膜时间常数只能实现有限的记忆长度，而异质 SNNs 则更加稳健，具有更好的泛化特性, 能够实现对于长期和短期记忆任务的良好的适应性. 值得一提的是，在本实验中我们仅使用了前馈的SNNs，也就是说，我们仅靠神经元的时间特性而不是循环的网络结构来实现工作记忆。这有效地说明了神经元的异质性在记忆中的重要作用。

As shown in Fig~\ref{fig:work_mem}, SNNs with heterogeneous neurons achieve significantly better memory performance than conventional SNNs, even in the case of non-optimizable synaptic weights. With increased task difficulty, the heterogeneous SNNs can maintain a reward of about $0.9$ at $25$ steps. In contrast, the performance of homogeneous SNNs sharply decreases as the memory length increases, even with adjustments to the connection weights. To ensure the robustness of our conclusions, we conducted a hyperparameter search for the membrane time constant of homogeneous SNNs, as depicted in the right of Fig~\ref{fig:work_mem}. For homogeneous SNNs, different membrane time constants can only achieve limited memory lengths, whereas heterogeneous SNNs demonstrate greater robustness and superior generalization, adapting well to both long-term and short-term memory tasks. It is worth noting that in this experiment, we only use feedforward SNNs, which means we rely solely on the time characteristics of neurons rather than recurrent structures to achieve working memory. This effectively demonstrates the critical role of neuron heterogeneity in memory.

\subsection{Neuronal Heterogeneity on Continuous Control Tasks}

% 我们在Brax中的连续机器人控制问题的基准上对于异质性的随机权重SNNs进行了评估, 并与连接权重可训练的同质化SNNs进行了对比. 对于所有的仿真环境, 我们比较了在使用不同的随机种子时, $5$次实验的累计奖励均值和标准差. 

We evaluate the performance of heterogeneous random-weight SNNs on a benchmark of continuous robot control problems in Google Brax library~\cite{brax2021github} and compare them with homogeneous SNNs with trainable connection weights. For all simulated environments, we reported the mean and standard deviation of the accumulated rewards across $5$ individual experiments with different random seeds.

% Ant、HalfCheetah、Hopper和Walker2d是Google Brax库中的强化学习环境，主要用于机器人运动控制的研究和实验。这些环境的主要目标都是通过学习策略，使机器人能够尽可能快地向前移动，并在某些环境中保持稳定的站立状态。对这些环境的具体介绍请您参考附录A.2. 

Ant (\textsc{Ant}), HalfCheetah (\textsc{Hc}), Hopper (\textsc{Hp}), and Walker2d (\textsc{Wal}) are reinforcement learning environments in the Brax, mainly used for research and experimentation in robot motion control. The primary objective of these environments is to enable robots to move forward as swiftly as possible while maintaining a stable standing position in specific scenarios. For detailed descriptions of these environments, please refer to Section~\ref{sec:brax}.

% ------------- Moved to Appendix ---------------
% % 所有环境都提供了关于机器人的位置，速度，角度以及关节角度的反馈信息，以便进行状态评估和决策。

% \texttt{Ant}, \texttt{HalfCheetah}, \texttt{Hopper}, and \texttt{Walker2d} are reinforcement learning environments in the Google Brax library~\cite{brax2021github}, mainly used for research and experimentation in robot motion control. The main goal of these environments is to use learning strategies to enable robots to move forward as quickly as possible and maintain stable standing positions in some environments. All environments provide feedback information about the robot's position, velocity, angle, and joint angles for state evaluation and decision-making.

% % Ant: 仿真四足机器人，奖励基于移动速度和生存时间。
% % HalfCheetah: 仿真二足机器人，奖励基于移动速度。
% % Hopper: 仿真单足机器人，奖励基于移动速度和生存时间。
% % Walker2d: 仿真二足行走机器人，奖励基于移动速度和生存时间。

% \begin{itemize}
%     \item \texttt{Ant}: Simulation of a quadruped robot with rewards based on movement speed and survival time.
%     \item \texttt{HalfCheetah}: Simulation of a bipedal robot resembling a half-cheetah with rewards based on movement speed.
%     \item \texttt{Hopper}: Simulation of a monopod robot with rewards based on movement speed and survival time.
%     \item \texttt{Walker2d}: Simulation of a bipedal walking robot with rewards based on movement speed and survival time.
% \end{itemize}

% 如表2所示, 与作为基线的传统的SNNs相比, 具有神经元异质性的SNNs在具有相同参数量的情况下, 能够表现出相同, 甚至更好的性能, 更具体的结果请参考附录A中的表S2. 随着可训练参数量的增长, 我们发现利用神经元异质性的随机网络的性能也会更高, 相反, 传统的SNNs的性能和可训练参数量之间的关系却并不显著. 这说明, 神经元异质性的随机网络可以更好地利用神经元的多样性，能够更好地捕捉数据的关系, 并提高网络的泛化性能. 相反, 传统的 SNNs 的神经元结构相对简单, 可能无法充分利用增加的可训练参数. 

\begin{figure}
    \centering
    \includegraphics[width=1.\linewidth]{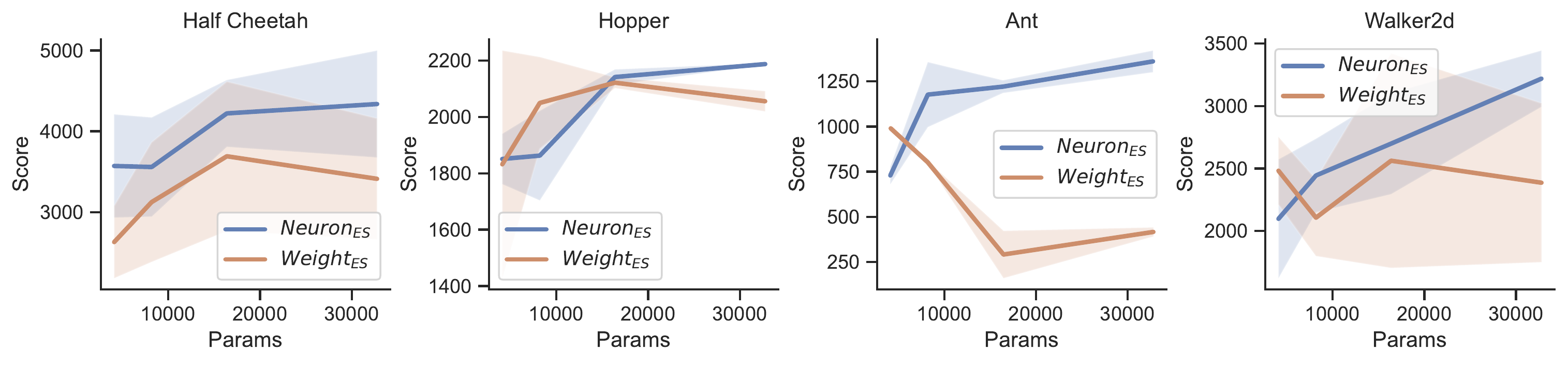}
    \caption{Comparison of rewards obtained from training neurons' properties and connection weights on continuous control tasks.}
    \label{diff_params}
\end{figure}

As shown in Fig.~\ref{diff_params}, SNNs with neuronal heterogeneity exhibit comparable or even better performance than conventional SNNs used as a baseline, with the same number of trainable parameters. As the number of trainable parameters increases, we observe that the performance of the random network with neuronal heterogeneity also improves. In contrast, the relationship between the performance of conventional SNNs and the number of trainable parameters is insignificant. This indicates that random networks with neuronal heterogeneity can better utilize the diversity of neurons, capture the relationships in the data, and improve the network's generalization performance. In contrast, the neuronal structure of conventional SNNs is relatively simple and may not fully utilize the increased number of trainable parameters.

\subsection{Ablation Study of Neuronal Properties}

% 我们设计了对于异质性神经网络的消融研究，旨在研究四个关键属性对网络性能的影响，包括膜时间常数、静息电位、阈值电压和膜电阻。实验通过分别移除这四个属性中的一个或几个，并观察网络性能的变化来评估各属性的重要性. 在所有的SNN中, 可训练的参数量被设置为16384. 更多神经元属性的组合消融实验可以在附录A.3的表S3中找到。

We conduct ablation studies on heterogeneous neural networks to investigate the impact of four key properties on network performance, including membrane time constant, resting potential, threshold voltage, and membrane resistance. The experiments involve removing one or more of these four properties and evaluating their importance by observing changes in network performance. In all SNNs, the number of trainable parameters is set to $16,384$.

\begin{table*}[htbp]
    \centering
    % $\checkmark$ 表示这一神经元的属性被设置为可训练的, 其他则表示神经元的属性被设置为固定的默认初始值. 
    \caption{Ablation study of neuronal properties. $\checkmark$ indicates that the neuron property is trainable, while others indicate that the neuron property is fixed to its default initial value.}
    \label{ablation}
    % \label{sample-table}
    \begin{tabular}{cccccccc}
        \toprule
        \multicolumn{4}{c}{\bf Neuronal Properties} & \multicolumn{4}{c}{\bf Reward (mean/stdev)}                                                                                                \\
        \cmidrule(lr){1-4} \cmidrule(lr){5-8}
        $\tau_m$                                    & $v_{th}$                                    & $v_{rest}$ & $R$        & \textsc{Hc}    & \textsc{Ant}   & \tt \textsc{Hp} & \textsc{Wal}   \\
        \hline
        \checkmark                                  & -                                           & -          & -          & $3967 \pm 171$ & $1180 \pm 200$ & $1796 \pm 9  $  & $2580 \pm 351$ \\
        -                                           & \checkmark                                  & -          & -          & $101  \pm 24 $ & $431  \pm 18 $ & $399  \pm 12 $  & $750  \pm 254$ \\
        -                                           & -                                           & \checkmark & -          & $3544 \pm 236$ & $680  \pm 19 $ & $1502 \pm 9  $  & $2286 \pm 549$ \\
        -                                           & -                                           & -          & \checkmark & $3264 \pm 528$ & $732  \pm 60 $ & $1654 \pm 29 $  & $2143 \pm 551$ \\
        \hline

        \checkmark                                  & \checkmark                                  & -          & -          & $3110 \pm 749$ & $1046 \pm 11 $ & $1982 \pm 154$  & $2866 \pm 402$ \\
        \checkmark                                  & -                                           & \checkmark & -          & $3887 \pm 637$ & $1160 \pm 215$ & $2016 \pm 94 $  & $2184 \pm 669$ \\
        \checkmark                                  & -                                           &            & \checkmark & $3460 \pm 331$ & $953  \pm 50 $ & $2031 \pm 13 $  & $3140 \pm 216$ \\
        -                                           & \checkmark                                  & \checkmark & -          & $3335 \pm 248$ & $765  \pm 172$ & $448  \pm 89 $  & $2880 \pm 174$ \\
        -                                           & \checkmark                                  & -          & \checkmark & $2679 \pm 406$ & $838  \pm 14 $ & $1771 \pm 12 $  & $2674 \pm 99 $ \\
        -                                           & -                                           & \checkmark & \checkmark & $3587 \pm 193$ & $457  \pm 413$ & $1902 \pm 9  $  & $2286 \pm 549$ \\
        \hline
        \checkmark                                  & \checkmark                                  & \checkmark & -          & $3924 \pm 342$ & $967 \pm 57$   & $1914 \pm 21$   & $2431 \pm 19$  \\
        \checkmark                                  & \checkmark                                  & -          & \checkmark & $3724 \pm 472$ & $1119 \pm 85$  & $1828 \pm 15$   & $2880 \pm 174$ \\
        \checkmark                                  & -                                           & \checkmark & \checkmark & $3982 \pm 319$ & $1276 \pm 18$  & $1943 \pm 34$   & $2779 \pm 73$  \\
        -                                           & \checkmark                                  & \checkmark & \checkmark & $3214 \pm 204$ & $978 \pm 21$   & $ 1790\pm 26$   & $2529 \pm 38$  \\
        \hline
        \checkmark                                  & \checkmark                                  & \checkmark & \checkmark & $4221 \pm 413$ & $1221 \pm 35$  & $2142 \pm 27$   & $2699 \pm 404$ \\
        \bottomrule
    \end{tabular}
\end{table*}

%             \checkmark   &  \checkmark  & -  & - &  $3110 \pm 749$ & $1046 \pm 11 $ & $1982 \pm 154$ & $2866 \pm 402$ \\  
%             \checkmark   &  -  & \checkmark  & - &  $3887 \pm 637$ & $1260 \pm 215$ & $2016 \pm 94 $ & $2184 \pm 669$ \\
%             \checkmark   &  -  &   & \checkmark  &  $3460 \pm 331$ & $953  \pm 50 $ & $2031 \pm 13 $ & $3140 \pm 216$ \\
%             -   &   \checkmark  & \checkmark  & - & $3335 \pm 248$ & $765  \pm 172$ & $448  \pm 89 $ & $2880 \pm 174$ \\
%             -   &   \checkmark  & -  & \checkmark & $2679 \pm 406$ & $838  \pm 14 $ & $1771 \pm 12 $ & $2674 \pm 99 $ \\
%             -   &  -  & \checkmark   & \checkmark & $3587 \pm 193$ & $457  \pm 413$ & $2102 \pm 9  $ & $2286 \pm 549$ \\
%             \hline 
%             \checkmark   & \checkmark & \checkmark& - & $ \pm $ & $ \pm $ & $ \pm $ & $ \pm $ \\  
%             \checkmark   &   \checkmark  & -  & \checkmark & $ \pm $ & $ \pm $ & $ \pm $ & $ \pm $ \\
%             \checkmark   &   -  & \checkmark  & \checkmark & $ \pm $ & $ \pm $ & $ \pm $ & $ \pm $ \\
%             -   &   \checkmark  & \checkmark  & \checkmark & $ \pm $ & $ \pm $ & $ \pm $ & $ \pm $ \\

\begin{figure}[!htbp]
    \centering
    \label{shapley}
    \includegraphics[width=.9\linewidth]{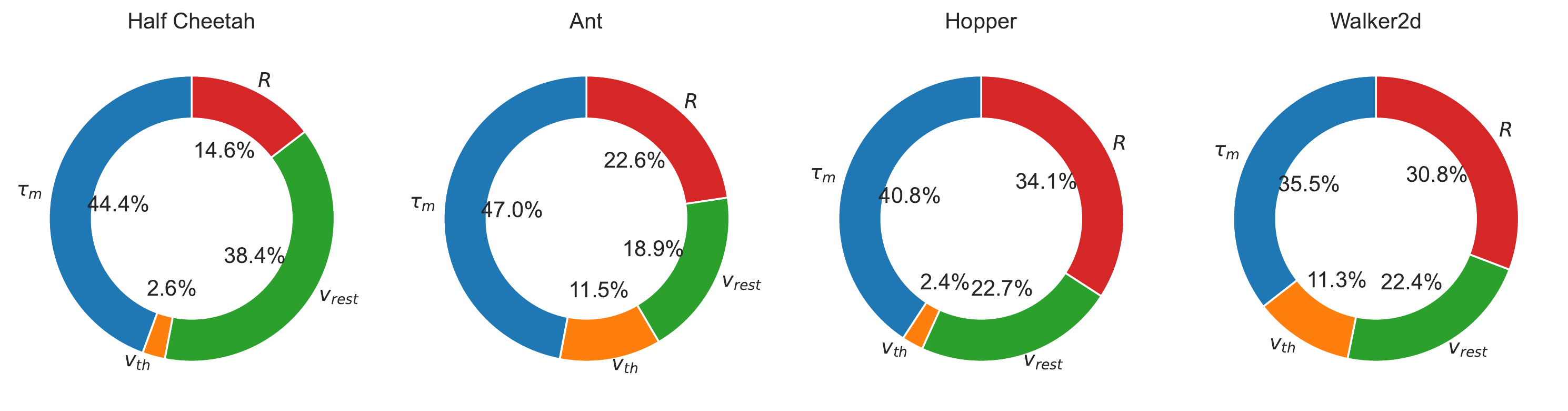}
    % 连续控制任务上, 优化权重/神经元参数时的性能对比. 手稿中Tab.1的可视化. 
    % 通过沙普利值对于异质性神经元不同属性的分析. 
    \caption{Analysis of different properties of heterogeneous neurons by Shapley values.}
\end{figure}

% 结果表明，膜时间常数对网络性能具有显著影响，而其他属性的影响相对较小。为了进一步定量地分析不同神经元的属性之间的相关性以及其对于模型性能的影响, 我们使用Shapely Value分析不同属性的平均预期边界贡献, 如图xxx所示. 这表明，在设计具有异质性神经元的网络时，膜时间常数是一个关键参数，需要特别关注以优化网络性能。
As shown in Tab.~\ref{ablation}, the results show that the membrane time constant significantly impacts the network performance, while other properties' influence is relatively minor. To further quantitatively analyze the correlations between different neuron attributes and their impact on model performance, we utilized Shapley Value analysis to assess the average expected boundary contributions of various attributes, as shown in Fig.~\ref{shapley}. This suggests that the membrane time constant is crucial when designing networks with heterogeneous neurons and requires special attention for optimizing network performance.

\subsection{Comparison of Membrane Time Constants Distributions and Experimental Data}

% 在本子小节中, 我们比较模型学习到的膜时间常数分布与一些生物学实验数据，以验证我们的异质性神经元是否与生物学现象相符合。我们在具有同质化初值的模型上对于训练好的膜时间常数的分布进行了可视化, 并与实验中观察到的小鼠V1脑区第四层刺状细胞以及人类颞叶中回的刺状细胞进行了对比, 如图xx所示. 

\begin{figure}
    \centering
    \includegraphics[width=1.\linewidth]{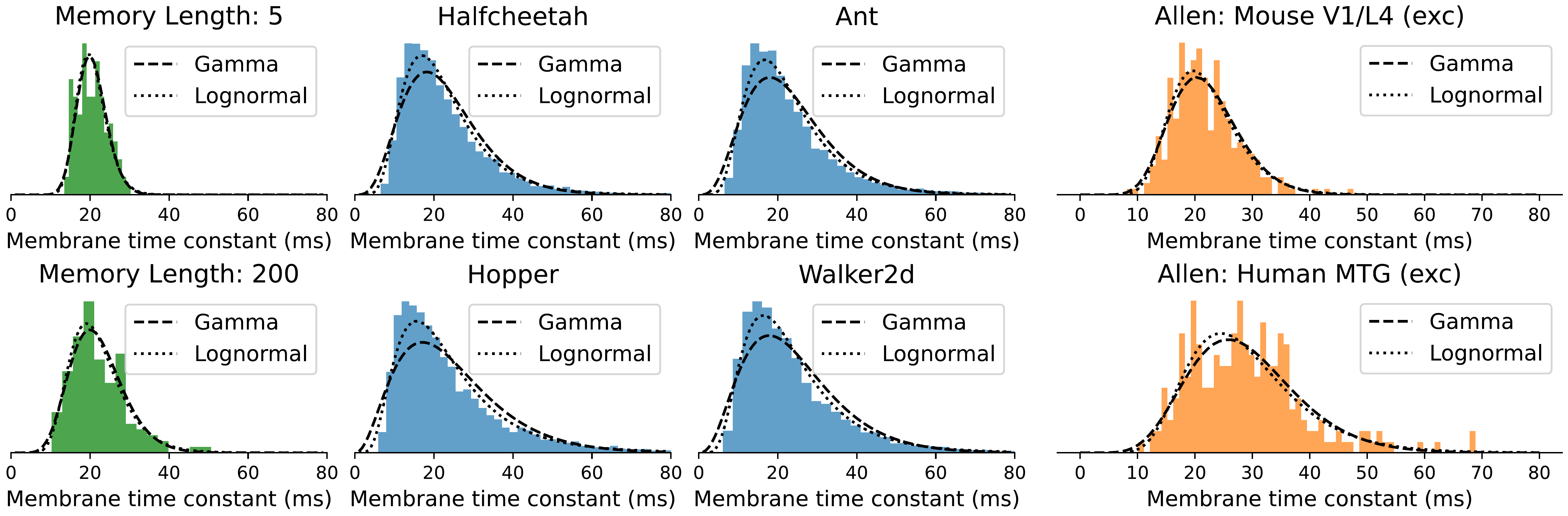}
    % 左: 强化任务中膜时间常数的分布. 右: 生物实验中膜时间常数的分布. 
    \caption{Left: Distribution of membrane time constants in reinforcement learning tasks.
        Right: Distribution of membrane time constants in biological experiments.}
    \label{fig:dist}
\end{figure}

In this subsection, we compare our model's membrane time constant distributions with some experimental biological data to verify whether our heterogeneous neurons are consistent with biological phenomena. We visualize the membrane time constant distribution learned from models with homogeneous initial values and compare it with that observed in layer $4$ spiny cells in mouse primary visual cortex~\cite{lein2007genome} and spiny cells in the temporal lobe gyrus of humans~\cite{hawrylycz2012anatomically}, as shown in Fig.~\ref{fig:dist}.

% 虽然我们的模型只是针对单一的强化学习任务, 但是模型中神经元的膜时间常数的分布和一些生物证据具有很强的相似性, 均呈现出长尾分布. 对于记忆长度任务, 我们发现对于短的延迟, 膜电位分布较为集中, 而在需要较长的记忆时间时, 膜时间常数就会增大以适应更长的延迟. 为了精确地衡量这一相似性, 我们使用了对数正态分布和Gamma分布对于数据进行了拟合, 结果如图xxx和表xxx所示. 

Although our model is only designed for a single task, the distribution of membrane time constants of the neurons in the model exhibits a substantial similarity to some biological evidence, which shows a long-tail distribution. For the \texttt{memory length} task, we find that for short delays, the membrane potential distribution is more concentrated, while for longer memory tasks, the membrane time constant increases to adapt to longer delays. To quantify this similarity, we fit the data using both lognormal distribution and gamma distribution, and the results are shown in Fig.~\ref{fig:dist} and Tab.~\ref{tab:dist}.

% 和生物神经元的相似性表明建立的模型能够在一定程度上模拟生物神经系统的特性。这种相似性可能为神经科学家提供更多关于神经元膜时间常数分布的理解, 同时这也表明异质性神经元在模拟生物神经元行为方面可能具有较高的有效性。这为使用这种模型进行进一步研究提供了信心。

The similarity with biological neurons suggests that the constructed model can simulate some characteristics of the biological nervous system to a certain extent. This similarity may provide neuroscientists with a better understanding of the distribution of membrane time constants of neurons and also indicates that heterogeneous neurons may have a higher effectiveness in simulating the behavior of biological neurons. This also provides confidence for further research using this model.

\begin{table*}[htbp]
    \centering
    % 这是不同强化学习任务以及生物实验中膜时间常数的拟合结果. 
    \caption{Fitting results of membrane time constants for reinforcement learning tasks and biological experiments. The location parameter is set to $0$.}
    \label{tab:dist}
    \begin{tabular}{cccccccc}
        \toprule
        \multicolumn{2}{c}{\bf Distributions} & \textsc{Hc} & \textsc{Ant} & \textsc{Hp} & \textsc{Wal} & \bf  Mouse V1 & \bf  Human MTG          \\
        \hline
        \multirow{2}{*}{gamma}                & shape       & $1.46$       & $1.85$      & $2.20$       & $1.85$        & $1.64$         & $3.18$ \\
                                              & scale       & $14.6$       & $11.7$      & $10.0$       & $11.8$        & $13.4$         & $9.11$ \\
        \hline
        \multirow{2}{*}{log normal}           & shape       & $0.26$       & $0.28$      & $0.30$       & $0.28$        & $0.27$         & $0.33$ \\
                                              & scale       & $20.6$       & $20.7$      & $20.9$       & $20.8$        & $21.1$         & $27.3$ \\
        \bottomrule
    \end{tabular}
\end{table*}

\subsection{Heterogeneous Spiking Neural Networks on Classification Tasks}

% 尽管模型针对的是单一的强化学习任务，但其神经元膜时间常数分布与生物学实验数据的相似性表明，该模型可能具有一定的通用性，能够在其他类似的任务中发挥作用。
Although the model is designed for a single reinforcement learning task, the similarity between its membrane time constant distribution and experimental biological data suggests that the model may have some universality and can be applied to other similar tasks.

% 在强化学任务上的结果验证了神经元异质性至关重要的作用, 即使在模型无法通过更新突触权重学习新的知识的情况下. 为了验证异质性SNNs在其他认知任务上的前景, 我们在在具有更高维度的图像分类任务上对其进行了验证. 我们在两个常用的图像分类数据集上进行了实验，分别是MNIST和FashionMNIST。其中MNIST包含10个数字类别（0-9），FashionMNIST则包含10个不同种类的服装。对于静态的图像数据, 需要较短的仿真时间, 因此对于这两个数据集, 仿真总时间被设置为$20$ ms, 即$4$个仿真步.

The results of the reinforcement learning task validate the crucial role of neural heterogeneity, even when the model cannot learn new knowledge through updating synaptic weights. We further test our model on higher-dimensional image classification tasks to verify the potential of heterogeneous SNNs in other tasks. Specifically, we test our model on two commonly used image datasets: MNIST~\cite{lecun1998mnist} and FashionMNIST~\cite{xiao2017fashion}. MNIST consists of 10-digit classes, while FashionMNIST contains $10$ different types of clothing. Since static image data requires a shorter simulation time, the total simulation time for both datasets was set to $4$ simulation steps.

\begin{table*}[htbp]
    \centering
    \caption{Comparison of heterogeneous SNNs and conventional SNNs on image classification tasks.}
    \label{tab:mnist}
    \begin{tabular}{cccccc}
        \toprule
        \bf Parameter   & \multicolumn{2}{c}{\bf MNIST} & \multicolumn{2}{c}{\bf FashionMNIST}                                 \\
        \cmidrule(lr){1-1} \cmidrule(lr){2-3} \cmidrule(lr){4-5}
        Neuron / Weight & Neuron                        & Weight                               & Neuron        & Weight        \\
        \hline
        4096  / 3970    & $\bf 80.96\%$                 & $78.66\%$                            & $\bf 76.35\%$ & $48.33\%$     \\
        8192  / 7940    & $85.64\%$                     & $\bf 88.69\%$                        & $\bf 77.64\%$ & $58.15\%$     \\
        16384 / 16674   & $88.89\%$                     & $\bf 90.59\%$                        & $79.31\%$     & $\bf 80.41\%$ \\
        32768 / 33348   & $90.84\%$                     & $\bf 91.29\%$                        & $80.90\%$     & $\bf 81.22\%$ \\
        \bottomrule
    \end{tabular}
\end{table*}

% 表xxx展示了具有不同参数量的异质性SNNs和同质性SNNs在图像分类任务上的测试准确率。当参数量较低时，异质性SNNs可能利用其内部的多样性来更好地捕捉输入数据的特征。与同质性SNNs相比，异质性SNNs的神经元具有不同的特性，这有助于网络学习更多样化的表示。随着参数量的增加，同质性SNNs可能逐渐克服了由于其结构简单性带来的局限。增加参数量可能使得网络具有更多的表达能力，从而在一定程度上弥补了同质性SNNs在处理复杂数据时的不足。因此，在参数量较低的情况下，异质性SNNs可能更适合处理复杂的任务，因为它们可以利用多样性来捕捉数据的特征。

Tab~.\ref{tab:mnist} shows the test accuracy of heterogeneous and homogeneous SNNs on the image classification task. When the number of parameters is low, heterogeneous SNNs may utilize their internal diversity to capture input data features better. Compared to homogeneous SNNs, the neurons in heterogeneous SNNs have different characteristics, which can help the network learn more diverse representations. As the number of parameters increases, homogeneous SNNs may gradually overcome the limitations caused by their simple structure. Increasing the number of parameters may give the network more expressive power, thereby partially compensating for the shortcomings of homogeneous SNNs in processing complex data. Herefore, in cases with low parameters, heterogeneous SNNs are more suitable for handling complex tasks because they can use diversity to capture the features of the data.

\begin{figure}
    \centering
    \includegraphics[width=.95\linewidth]{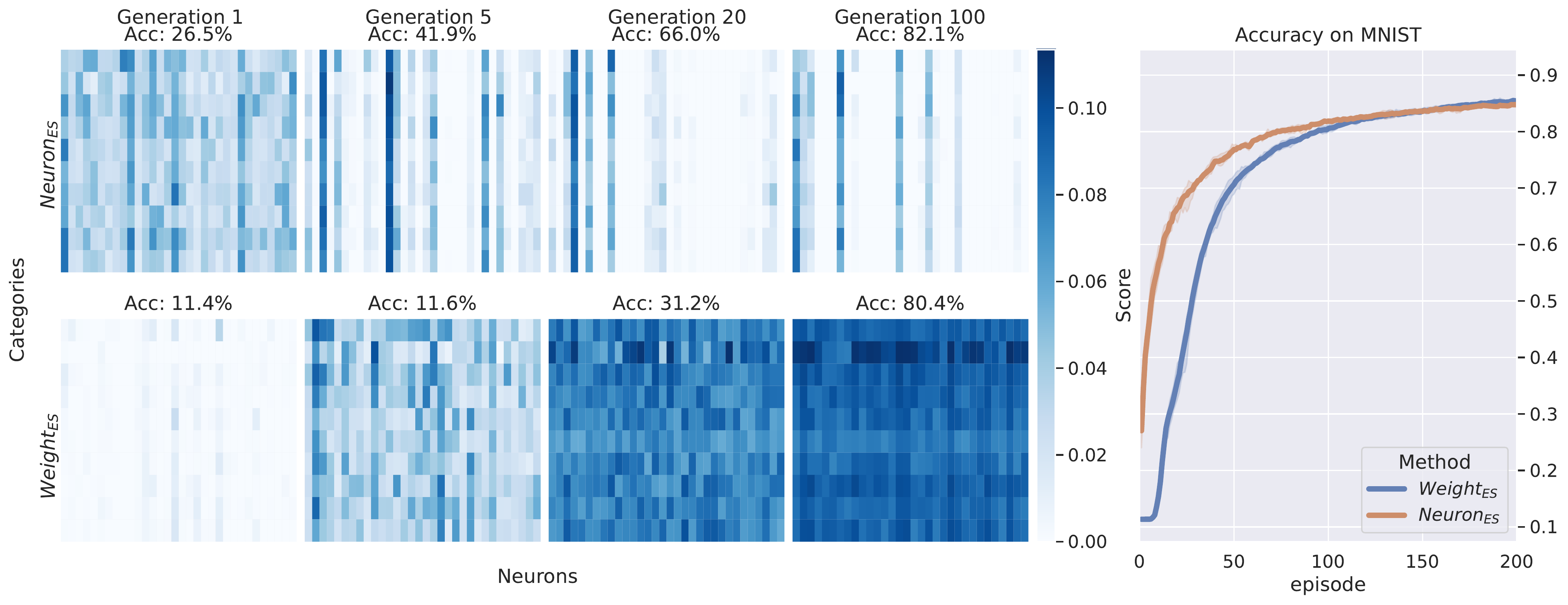}
    % MNIST数据集上对不同的类别, 随机选取的$32$个神经元在不同代的脉冲频率。
    \caption{Spike frequency of randomly selected $32$ neurons in different generations for different categories on the MNIST dataset.}
    \label{fig:fire_rate}
\end{figure}

% 除了与模型准确率相关的结论外，我们还发现了一个有趣的现象：与同质性的SNNs相比，异质性网络中神经元的活动模型具有显著的差异。如图xxx所示，我们随机选取了一些神经元，并可视化了在输入为不同类别时，神经元的平均脉冲率。我们发现在训练的初始阶段，异质性SNNs中，几乎所有的神经元都有脉冲活动，同时模型能够更快地收敛。随着训练的进行，只有少数的神经元仍保持较高的脉冲率，而大多数神经元则静默了。这些活跃的神经元能够通过不同的脉冲率，实现对于不同类别刺激的高效响应。相较之下，同质性SNNs在权重优化过程中，神经元的脉冲率随着训练的进行不断增大。尽管模型最后也能达到较高的性能，但其代价是神经元更高的平均脉冲率。

In addition to the conclusions related to accuracy, we have uncovered an intriguing phenomenon: notable distinctions in the activity patterns of neurons in heterogeneous networks compared to homogeneous SNNs. As depicted in Fig.~\ref{fig:fire_rate}, we randomly select a subset of neurons and visualize their average spiking rates for different input stimuli representing different categories. It is observed that during the initial stages of training, nearly all neurons in the heterogeneous SNNs exhibit spiking activity, resulting in faster convergence of the model. As training progressed, only a minority of neurons maintain high firing rates, while most remain silent. These active neurons can efficiently respond to stimuli from different categories through their special firing rates. In contrast, homogeneous SNNs exhibit an increasing trend in neuron firing rates during weight optimization. Although homogeneous models eventually achieve high performance, it comes at the cost of higher average firing rates of neurons.

\subsection{Discussion on Backpropagation and Evolutionary Strategies}

% 正如在3.2节讨论的, 反向传播在应用于不稳定系统的动态系统时候, 可能会导致发散的结果. 而LIF神经元由于不可微分, 需要引入代理梯度函数来建立近似的反向传播链路. 但是这种估计会导致梯度估计的不准确. 而随着任务序列长度的增加, 由此造成的梯度估计的方差会导致系统更加不稳定, 甚至发散. 因此我们引入无梯度的演化算法对于异质性SNNs进行优化. 虽然我们失去了一个维度的效率因素, 但是我们可以获得更稳定的对于优化方向的估计. % 为了证实上述理论, 我们在一些经典的控制任务上对比了基于梯度的优化方法和基于黑箱优化的方法在同质性和异质性SNNs上的性能. 

As discussed in Section~\ref{sec:dynamic_system}, BP may lead to divergent results when applied to dynamic systems of unstable nature. In the case of LIF neurons, which are non-differentiable, surrogate gradient functions need to be introduced to establish an approximate BP pathway. However, this estimation may cause inaccurate gradient estimates. As the length of the task sequence increases, the resulting variance of the gradient estimation can lead to a more unstable system. Therefore, we use gradient-free ES for optimizing heterogeneous SNNs. Although we lose a dimension of the efficiency factor, we can obtain more stable estimates of the optimization direction. To verify the above theory, we compared the performance of gradient-based optimization methods~\cite{williams1992simple} and black-box optimization methods on homogeneous and heterogeneous SNNs in some classical control tasks.

% --------------------- Moved to Appendix -------------
% 在使用BPTT进行优化时, 优化器是Adam~\cite{kingma2014adam}, 学习率$lr_{adam}=3 \times 10^{-2}$. 为了解决脉冲神经元不可微分的问题, 与~\cite{wu2018spatio}相同的代理梯度函数被使用, 以建立网络的反向传播通路.

% When using BPTT, the optimizer was Adam~\cite{kingma2014adam}, with a learning rate $lr_{bp} = 3 \times 10^{-2}$.  To address the non-differentiability issue of spiking neurons, the same surrogate gradient function as in~\cite{wu2018spatio} was used to establish the backward propagation path of the SNNs.

\begin{figure}[htbp]
    \centering
    \includegraphics[width=.95\linewidth]{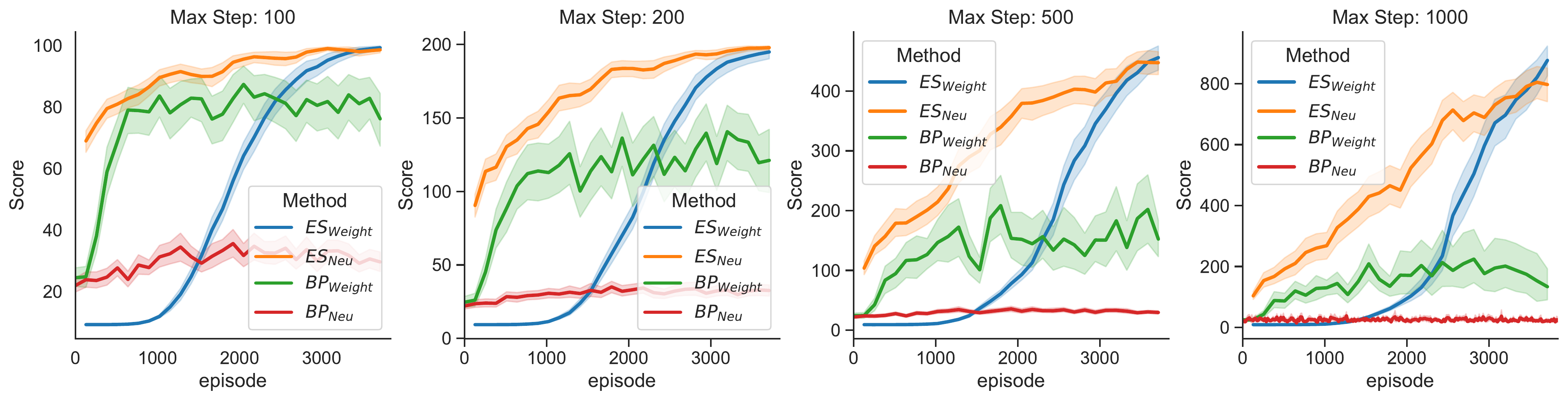}
    % 在CartPole-v1环境上不同优化方法在同质性/异质性SNNs上的性能.
    \caption{Performance of different optimization methods on homogeneous/heterogeneous SNNs on CartPole-v1 environment.}
    \label{fig:curve}
\end{figure}

% 如图xxx所示, 我们在CartPole-v1任务上研究了不同的序列长度对于模型性能的影响. 我们分别测试了最大的步数为100, 200, 500和1000时, 模型的性能, 对不同的设置, 我们使用了不同的种子重复了100次. 对于ES而言, 种群数量被设置为128, 迭代次数被设置为30, 也就是说, 一共和环境交互3840个回合. 横坐标表示所有的个体与环境交互的总的回合数量. 而纵坐标则表示智能体的累积奖励, 其意义是在倒立摆的极点角度小于$12\dgree$的步数. 

As shown in Fig~\ref{fig:curve}, we study the effect of different sequence lengths on the model's performance in the CartPole-v1 task. We tested the model's performance with a maximum number of steps set to $100$, $200$, $500$, and $1000$, and repeated each experiment $100$ times with different random seeds. For ES, the population size is set to 128, and the number of iterations was set to $30$, which means a total of $3,840$ episodes of interaction with the environment. The y-axis represents the cumulative reward of the agent, which means the number of steps where the pole angle is less than $12^{\circ}$.

% 对于同质性SNNs, 基于BP的方法比ES能够更快地收敛. 但是随着序列的变长, 基于BP的SNNs难以收敛到最优的值, 而ES却能稳健地收敛到具有更高奖励的状态. 同时, 基于BP的方法很难处理权重不可优化的情况, 相反, ES则能够更高效地和环境进行交互, 在较少的回合就能取得超过基于BP的结果. 这说明, ES在搜索解空间时能够更好地利用这种神经元的多样性, 从而在某些情况下找到更优的解. 同时, ES通过直接与环境进行交互来评估和优化策略，这种方法可能更接近于生物神经系统中的学习过程. 
For homogeneous SNNs, the BP-based method converges faster than ES. However, as the sequence length increases, BP-based SNNs struggle to converge to the optimal value, while ES can robustly converge higher rewards. Furthermore, BP-based methods cannot handle untrainable weights, whereas ES can efficiently interact with the environment and achieve better results than BP-based methods in fewer episodes. This suggests that ES can better exploit the diversity of neural elements when searching the solution space and thus find better solutions in some cases. Additionally, ES evaluates and optimizes policies by directly interacting with the environment, which may be closer to the learning process in biological neural systems.

\section{Discussion}

In this study, we investigate the role of neuronal heterogeneity in SNNs, with a particular focus on optimizing neuronal properties without modifying the connection weights. Our research reveals the challenges faced by BP-based methods in optimizing spiking neurons, especially in dealing with long sequence tasks. As an alternative approach, we find that evolutionary strategies perform better in optimizing neuronal parameters in random networks. Through experiments on tasks such as working memory, continuous control, and image recognition, we demonstrate that optimizing neuronal properties can achieve comparable or even superior performance compared to optimizing connection weights. This highlights the importance of neuronal heterogeneity in network performance, particularly in tasks with rich temporal structures. We emphasize the role of membrane time constants in neuronal heterogeneity, and our model distribution reflects observations from biological experiments. This work contributes to the understanding of neuronal heterogeneity in both biological and SNNs, providing insights for future research in biologically-inspired neural networks.

While our study contributes to understanding neuronal heterogeneity in biological and SNNs, there are limitations that should be acknowledged. We used simplified models of spiking neurons, and real biological neurons exhibit more complex behaviors influenced by various physiological factors, all of which contribute to neuronal heterogeneity. Therefore, the behavior of our model may not fully capture the behavior of biological neurons.

In our future work, we plan to address these limitations by studying more complex neuronal models, integrating more biological observations, exploring novel optimization methods, and testing in broader tasks and domains. By delving deeper into these areas, we aim to advance our understanding of neural networks and enhance their capabilities.

\section{Network Settings and Training Details}

\subsection{Spiking Neurons: System Stability Analysis}

% 脉冲神经元可以被视为一个迭代的动态系统. 而在使用梯度优化进行优化时, 正如正文中所说, 其重置后膜电位在时间维度上的梯度可以被表示为:
The spiking neuron can be regarded as an iterative dynamic system. When using gradient optimization for optimization, as mentioned in the text, the gradient of the reset membrane potential over time can be represented as follows:

\begin{equation}
    \frac{\partial v(t)}{\partial v(t - \Delta t)} = [1 - s(t) + (v_{rest} - u(t)) g^\prime (u(t) - v_{th})](1 - \frac{\Delta t
    }{\tau_m})
    \label{recurisve_v1}
\end{equation}

% 在式1中, $u(t)$ 表示重置前膜电位, 而其常常是无界的, 我们可视化了在使用CartPole-v1任务上, 使用BPTT训练SNNs连接权重时, 神经元的膜时间常数. 神经元参数使用和Tab.S1中相同的设置. 

In Eq~\ref{recurisve_v1}, $u(t)$ represents the membrane potential before the reset, which is unbounded. We visualize the membrane time constants of neurons when training SNNs with BPTT on the \texttt{CartPole-v1}. The neuron parameters are the same as those in Tab.~\ref{params}.

\begin{figure}[htbp]
    \centering
    \includegraphics[width=1.\linewidth]{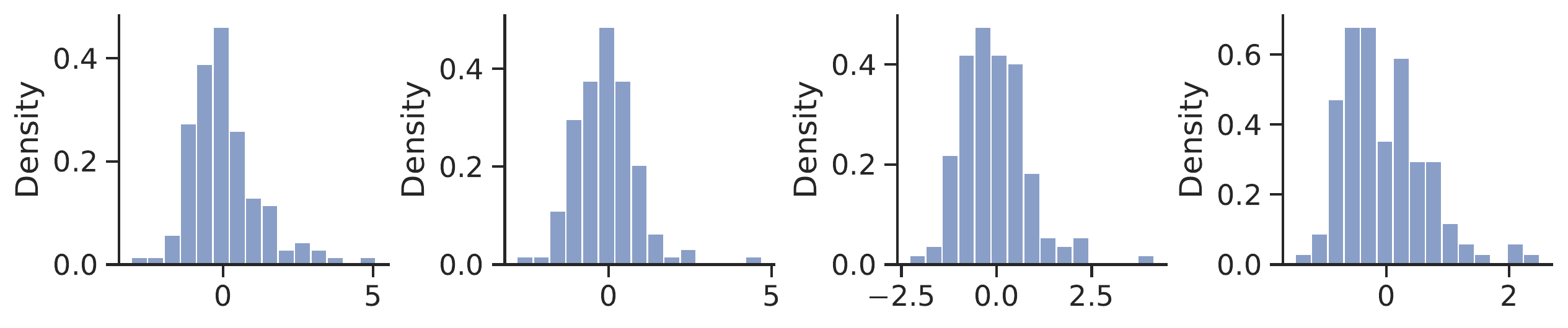}
    % CartPole中, 4个不同的回合中, 膜电位的分布.
    \caption{The distribution of membrane potentials in four different episodes of the \texttt{CartPole-v1}.}
    \label{fig:mem_dist}
\end{figure}

% 如图S1所示, 在实践中, $u(t)$ 是无界的, 而且能达到约$5$. 这使得$\frac{\partial v(t)}{\partial v(t - \Delta t)}$的大小可能大于$1$, 进而导致系统发散. 而且在Eq.1中存在 $1-s(t)$, 因此只要Eq.1中$(v_{rest} - u(t)) g^\prime (u(t) - v_{th}) \geq 0$ 同时神经元没有兴奋, 这就有使得系统发散的风险. 

As shown in Fig~\ref{fig:mem_dist}, in practice, $u(t)$ is unbounded and can reach approximately $5$. This can result in the magnitude of $\frac{\partial v(t)}{\partial v(t - \Delta t)}$ being greater than $1$, leading to system divergence. Additionally, in Eq.~\ref{recurisve_v1}, there is a term $1-s(t)$, which means that as long as $(v_{rest} - u(t)) g^\prime (u(t) - v_{th}) \geq 0$ and the neuron is not firing, there is a risk of system divergence.

% --------------- removed ---------------
% 由于Heaviside 函数不可微分, 因此需要使用代理梯度~\cite{bohte2011error}对于$g(\cdot)$的梯度进行近似. 一种常用的代理梯度函数可以被表示为:

Due to the non-differentiability of the Heaviside function, it is necessary to approximate the gradient of $g(\cdot)$ using a surrogate gradient~\cite{bohte2011error}. One commonly used proxy gradient function can be represented as follows:

\begin{equation}
    g^\prime (u(t) - v_{th}) = \frac{1}{v_{th}^2} \max (0, v_{th} - \left| u(t) - v_{th} \right|)
\end{equation}

% 在实际应用中, 有两种策略用于对脉冲神经元进行优化. 一种是如式1所示的, 考虑脉冲和重置对于神经元的影响. 另一种则是忽略重置对于梯度的影响, 可以被表示为:
In practical applications, there are two strategies for optimizing spiking neurons. One approach considers the impact of spikes and resets on the neuron, as shown in Eq.~\ref{recurisve_v1}. The other approach ignores the influence of resets on gradients and can be represented as follows:

\begin{equation}
    \frac{\partial v(t)}{\partial v(t - \Delta t)} = [1 - s(t)](1 - \frac{\Delta t}{\tau_m})
    \label{recurisve_v2}
\end{equation}

% Eq.2忽略了$\frac{\partial v(t))}{\partial s(t)}$, 这使得$ \frac{\partial v(t)}{\partial v(t - \Delta t)} < 1$, 保证了系统的稳定性. 但是这样使得时间维度的梯度随着序列长度呈现指数衰减, 使得网络难以建立鲁棒的时序依赖, 限制了模型处理具有强时间相关性的任务的能力. 特别地, 对于神经元属性而言, 这样则会将$v_{rest}$ 和 $v_{th}$ 从时间梯度链路上移除, 进一步限制了异质性SNNs的潜力. 

Eq.~\ref{recurisve_v2} neglects the term $\frac{\partial v(t)}{\partial s(t)}$, which ensures that $\left|\frac{\partial v(t)}{\partial v(t - \Delta t)}\right| < 1$, thereby maintaining system stability. However, this leads to the exponential decay of the temporal gradients with increasing sequence length, making it challenging for the network to establish robust temporal dependencies and limiting its ability to handle tasks with strong temporal correlations. Specifically, for neuron properties, this also removes $v_{rest}$ and $v_{th}$ from the temporal gradient path, further restricting the potential of heterogeneous SNNs.

\subsection{Network Structure and Hyperparameters}

\begin{table*}[htbp]
    \centering
    \caption{Initial parameters of the spiking neurons.}
    \label{params}
    \begin{tabular}{ccc}
        \toprule
        \bf Parameter & \bf Value                & \bf Description            \\
        \hline
        $\Delta t$    & $5$ ms                   & Simulation time step       \\
        $\tau_m$      & $20$ ms                  & Membrane time constant     \\
        $v_{th}$      & $0.5$ V                  & Membrane threshold         \\
        $v_{rest}$    & $0$ mV                   & Resting \& Reset potential \\
        $R$           & $5 \times 10^{7} \Omega$ & Membrane Resistance        \\
        \bottomrule
    \end{tabular}
\end{table*}

% 在所有的实验中, 网络结构都被设置为2层的感知机, 并通过调整隐藏层中神经元的数量实现不同的可训练参数. 所使用的LIF模型如Eq.xxx所示, 并使用和xxx相同的方法调控神经元的脉冲. 在训练连接权重时, 神经元的参数设置如表xxx所示. 在优化异质性神经元时, 突触权重将被固定, 神经元的$\tau_m, v_{th}, v_{rest}, R$ 会被设置为可训练参数, 并且初始值和训练连接权重时候相同. 而对于$\tau_m$, 则通过$k(\tau_m) = \frac{1}{1 + e^{-\tau_m}}$ 对膜时间常数进行间接优化. 对于所有的情况, 连接权重都使用LeCun初始化~\cite{klambauer2017self}. PGPE被作为主要的优化算法, 种群数量被设置为$256$, 如无特殊说明, 训练的代数为$1000$, 最后一代的所有个体的平均的奖励和标准差将被报告. 中心学习率$lr_{center}$ 被设置为 $0.15$, 高斯分布标准差的学习率$lr_{std}$ 被设置为 $0.1$, 初始的标准差$\sigma_0 = 0.1$. 

We conduct experiments using the Jax framework~\cite{jax2018github} with NVIDIA A100 40G GPU. In all experiments, the network architecture is set as a 2-layer perception, with varying trainable parameters achieved by adjusting the number of neurons in the hidden layer. The LIF model is shown in Eq.~\ref{lif}, and the same method as~\cite{zhao2022backeisnn} controls the neuron's spiking. When training the connection weights, the parameters of the neurons are set as shown in Tab.~\ref{params}. When optimizing heterogeneous neurons, synaptic weights were fixed, and neuron parameters, including $\tau_m, v_{th}, v_{rest}, R$, are set as trainable parameters with the same initial values as when training connection weights. For $\tau_m$, an indirect optimization of the membrane time constant is performed using $k(\tau_m) = \frac{1}{1 + e^{-\tau_m}}$.

For all cases, connection weights are initialized using LeCun initialization~\cite{klambauer2017self}. PGPE~\cite{sehnkeParameterexploringPolicyGradients2010} is used as the optimization algorithm, with a population size of $256$, and unless otherwise specified, trained for $1000$ generations. The average reward and standard deviation of all individuals in the last generation are reported. The center learning rate $lr_{center}$ is set to $0.15$, the learning rate for the standard deviation of the Gaussian distribution $lr_{std}$ is set to $0.1$, and the initial standard deviation $\sigma_0$ are set to $0.1$.

% 在使用BPTT进行对比试验时, 优化器是Adam~\cite{kingma2014adam}, 学习率$lr_{adam}=3 \times 10^{-2}$. 为了解决脉冲神经元不可微分的问题, 与~\cite{wu2018spatio}相同的代理梯度函数被使用, 以建立网络的反向传播通路.

For comparison experiments using BPTT, the optimizer is Adam~\cite{kingma2014adam}, with a learning rate $lr_{bp} = 3 \times 10^{-2}$.  To address the non-differentiability issue of spiking neurons, the same surrogate gradient function as in~\cite{wu2018spatio} is used to establish the backward propagation path of the SNNs.

% 对于反向传播而言, 有很多参数会影响优化过程的稳定性, 例如学习率, 以及对于SNNs而言独有的代理梯度函数的形状及其参数. 因此为了说明文中结论的鲁棒性, 我们在CartPole-v1任务上, 对使用反向传播优化时的包括学习率以及代理梯度等超参数进行了搜索. 

In the context of backpropagation, several parameters can influence the stability of the optimization process, such as the learning rate and, uniquely for SNNs, the shape and parameters of the surrogate gradient function. Therefore, to demonstrate the robustness of the conclusions in this paper, we conducted a search for hyperparameters, including learning rates and surrogate gradient-related parameters, when optimizing on the CartPole-v1 task using backpropagation.

\begin{table}[htbp]
    \centering
    \caption{Hyperparametric search for learning rate $lr_{bp}$ on CartPole-v1 task.}
    \begin{tabular}{ccccccc}
        \hline
                                       & \bf Max step & $1 \times 10^{-3}$ & $1 \times 10^{-2}$ & $3 \times 10^{-2}$ & $0.1$ & $0.3$ \\
        \hline
        \multirow{4}{*}{Weight$_{BP}$} & 100          & 55.7               & 84.3               & 81.2               & 90.2  & 39.9  \\
                                       & 200          & 64.7               & 124.6              & 127.3              & 146.1 & 94.6  \\
                                       & 500          & 52.3               & 161.5              & 183.7              & 201.1 & 34.6  \\
                                       & 1000         & 52.9               & 114.8              & 190.7              & 228.3 & 40.6  \\
        \hline
        \multirow{4}{*}{Neuron$_{BP}$} & 100          & 46.9               & 55.4               & 55.3               & 62.2  & 50.4  \\
                                       & 200          & 55.42              & 63.9               & 62.3               & 63.1  & 57.6  \\
                                       & 500          & 60.0               & 73.4               & 63.5               & 74.4  & 61.0  \\
                                       & 1000         & 50.4               & 73.9               & 76.4               & 80.5  & 62.9  \\
        \hline
    \end{tabular}
    \label{search_lr}
\end{table}

\begin{table}[htbp]
    \centering
    \caption{Hyperparametric search for surrogate function on CartPole-v1 task.}
    \resizebox{\linewidth}{!}{
        \begin{tabular}{cccccc|cccc}
            \hline
            \makecell{Surrogate                                                                                               \\ Function}   &          & \multicolumn{3}{c}{$g(x) = \frac{1}{\pi} \arctan(\frac{\pi}{2}\alpha x) + \frac{1}{2}$}  &  & \multicolumn{4}{c}{$g(x) = \mathrm{NonzeroSign}(x) \log (|\alpha x| + 1)$}                                                    \\
            \hline
                                           & \bf Max step & $a=0.5$ & $a=1$ & $a=2$ & $a=4$ & $a=0.5$ & $a=1$ & $a=2$ & $a=4$ \\
            \hline
            \multirow{4}{*}{Weight$_{BP}$} & 100          & 87.1    & 87.3  & 81.2  & 85.7  & 79.8    & 84.7  & 85.7  & 81.3  \\
                                           & 200          & 120.4   & 157.6 & 127.3 & 146.7 & 117.5   & 124.6 & 119.3 & 102.6 \\
                                           & 500          & 201.3   & 232.6 & 183.7 & 177.7 & 166.7   & 159.3 & 172.8 & 161.3 \\
                                           & 1000         & 197.6   & 259.1 & 190.7 & 201.4 & 285.2   & 217.5 & 228.7 & 217.6 \\
            \hline
            \multirow{4}{*}{Neuron$_{BP}$} & 100          & 54.2    & 57.8  & 55.3  & 64.5  & 41.3    & 46.8  & 57.8  & 65.3  \\
                                           & 200          & 56.4    & 70.8  & 62.3  & 70.1  & 45.0    & 44.2  & 73.0  & 82.8  \\
                                           & 500          & 65.9    & 65.2  & 63.5  & 63.2  & 51.3    & 55.3  & 74.4  & 81.2  \\
                                           & 1000         & 70.8    & 68.2  & 76.4  & 69.1  & 51.2    & 48.3  & 75.8  & 74.3  \\
            \hline
        \end{tabular}
    }
    \label{search_sg}
\end{table}

% 如表Tab~\ref{search_lr, search_sg}所示, 在 Cartpole 任务中，代梯度函数和学习率等超参数的设置与性能之间的关系并不明显，这证明了手稿中结论的稳健性。特别是对于异构 SNN，用 BP 对其进行优化是一项挑战, 并且其性能显著地弱于基于演化的优化策略。

As shown in Tab.~\ref{search_lr} and Tab.~\ref{search_sg}, the relationship between hyperparameters such as surrogate gradient functions and learning rates and performance in the CartPole task is not very pronounced. This observation further underscores the robustness of the conclusions in the manuscript. Particularly, for heterogeneous SNNs, optimizing them with backpropagation presents a challenge, and their performance is significantly inferior to evolution-based optimization strategies.

\subsection{Introduction to the Continuous Control Environments}
\label{sec:brax}
% 所有环境都提供了关于机器人的位置，速度，角度以及关节角度的反馈信息，以便进行状态评估和决策。

Ant (\textsc{Ant}), HalfCheetah (\textsc{Hc}), Hopper (\textsc{Hp}), and Walker2d (\textsc{Wal})  are reinforcement learning environments in the Google Brax library~\cite{brax2021github}, mainly used for research and experimentation in robot motion control. The main goal of these environments is to use learning strategies to enable robots to move forward as quickly as possible and maintain stable standing positions in some environments. All environments provide feedback information about the robot's position, velocity, angle, and joint angles for state evaluation and decision-making.

% Ant: 仿真四足机器人，奖励基于移动速度和生存时间。
% HalfCheetah: 仿真二足机器人，奖励基于移动速度。
% Hopper: 仿真单足机器人，奖励基于移动速度和生存时间。
% Walker2d: 仿真二足行走机器人，奖励基于移动速度和生存时间。

\begin{itemize}
    \item \textsc{Ant}: Simulation of a quadruped robot with rewards based on movement speed and survival time.
    \item \textsc{Hc}: Simulation of a bipedal robot resembling a half-cheetah with rewards based on movement speed.
    \item \textsc{Hp}: Simulation of a monopod robot with rewards based on movement speed and survival time.
    \item \textsc{Wal}: Simulation of a bipedal walking robot with rewards based on movement speed and survival time.
\end{itemize}

\section*{Acknowledgement}
This work was supported by the National Key Research and Development Program (Grant No. 2020AAA0104305), and the Strategic Priority Research Program of the Chinese Academy of Sciences (Grant No. XDB32070100).

\bibliography{refs}
\bibliographystyle{unsrt}

\end{document}